\pdfoutput=1

\documentclass[11pt]{article}

\usepackage{authblk}
\usepackage{emnlp_template/emnlp2021}

\usepackage{times}
\usepackage{latexsym}

\usepackage[T1]{fontenc}

\usepackage[utf8]{inputenc}

\usepackage{microtype}

\usepackage{subfiles}

\usepackage{booktabs}
\usepackage{tabularx}
\usepackage{multirow}
\usepackage{siunitx}
\usepackage{arydshln}
\usepackage{array}
\usepackage{lipsum}

\usepackage{adjustbox}
\usepackage{array}
\usepackage{xcolor}

\newcolumntype{Y}{>{\centering\arraybackslash}X}
\newcolumntype{R}{>{\raggedleft\arraybackslash}X}

\newcommand{\raisemath}[1]{\mathpalette{\raisemeth{#1}}}
\newcommand{\raisemeth}[3]{\raisebox{#1}{$#2#3$}}
\newcommand{\stddev}{\raisemath{-3pt}}{}

\newcommand\blfootnote[1]{%
  \begingroup
  \renewcommand\thefootnote{}\footnote{#1}%
  \addtocounter{footnote}{-1}%
  \endgroup
}

\title{Separating Retention from Extraction in \\ the Evaluation of End-to-end Relation Extraction}
\author[1]{Bruno Taill\'{e}}
\author[1]{Vincent Guigue}
\author[2]{Geoffrey Scoutheeten}
\author[1,3]{Patrick Gallinari}

\affil[1]{Sorbonne Universit\'{e}, CNRS, Laboratoire d'Informatique de Paris 6, LIP6}
\affil[ ]{\textsuperscript{2}BNP Paribas \hspace{2cm} \textsuperscript{3}Criteo AI Lab}
\affil[ ]{\texttt{\{bruno.taille, vincent.guigue, patrick.gallinari\}@lip6.fr}}
\affil[ ]{\texttt{geoffrey.scoutheeten@bnpparibas.com}}

\begin{document}
\maketitle
\begin{abstract}

State-of-the-art NLP models can adopt shallow heuristics that limit their generalization capability \cite{mccoy-etal-2019-right}.
Such heuristics include lexical overlap with the training set in Named-Entity Recognition \cite{Taille2020ContextualizedGeneralization} and \textit{Event} or \textit{Type} heuristics in Relation Extraction \cite{rosenman-etal-2020-exposing}.
In the more realistic end-to-end RE setting, we can expect yet another heuristic: the mere retention of training relation triples.
In this paper we propose several experiments confirming that retention of known facts is a key factor of performance on standard benchmarks.
Furthermore, one experiment suggests that a pipeline model able to use intermediate type representations is less prone to over-rely on retention.

\end{abstract}

\section{Introduction}
\label{sec:introduction}
\blfootnote{Code for reproducing our evaluation settings is available at \href{https://github.com/btaille/retex}{github.com/btaille/retex}}
Information Extraction (IE) aims at converting the information expressed in a text into a predefined structured format of knowledge.
This global goal has been divided into subtasks easier to perform automatically and evaluate.
Hence, Named Entity Recognition (NER) and Relation Extraction (RE) are two key IE tasks among others such as Coreference Resolution (CR), Entity Linking or Event Extraction.
Traditionally performed as a pipeline \cite{Bach2007AExtraction}, these two tasks can be tackled jointly in order to model their interdependency, alleviate error propagation and obtain a more realistic evaluation setting \cite{Roth2002ProbabilisticRecognition, Li2014IncrementalRelations}.

Following the general trend in Natural Language Processing (NLP), the recent quantitative improvements reported on Entity and Relation Extraction benchmarks are at least partly explained by the use of larger and larger pretrained Language Models (LMs) such as BERT \cite{devlin-etal-2019-bert} to obtain contextual word representations.
Concurrently, there is a realization that new evaluation protocols are necessary to better understand the strengths and shortcomings of the obtained neural network models, beyond a single holistic metric on an hold-out test set \cite{ribeiro-etal-2020-beyond}.

In particular, generalisation to unseen data is a key factor in the evaluation of deep neural networks.
It is all the more important in IE tasks that revolve around the extraction of mentions: small spans of words that are likely to occur in both the evaluation and training datasets.
This lexical overlap has been shown to be correlated to neural networks performance in NER \cite{Augenstein2017GeneralisationAnalysis, Taille2020ContextualizedGeneralization}.
For pipeline RE, \citet{rosenman-etal-2020-exposing} and \citet{peng-etal-2020-learning} expose shallow heuristics in neural models: relying too much on the type of the candidate arguments or on the presence of specific triggers in their contexts.

In end-to-end Relation Extraction, we can expect that these NER and RE heuristics are combined.
In this work, we argue that current evaluation benchmarks measure both the desired ability to extract information contained in a text but also the capacity of the model to simply retain labeled (head, predicate, tail) triples during training.
And when the model is evaluated on a sentence expressing a relation seen during training, it is hard to disentangle which of these two behaviours is predominant.
However, we can hypothesize that the model can simply retrieve previously seen information acting like a mere compressed form of knowledge base probed with a relevant query.
Thus, testing on too much examples with seen triples can lead to overestimate the generalizability of a model.

Even without labeled data, LMs are able to learn some relations between words that can be probed with cloze sentences where an argument is masked  \cite{petroni-etal-2019-language}.
This raises the additional question of lexical overlap with the orders of magnitude larger unlabeled LM pretraining corpora that will remain out of scope of this paper.


\section{Datasets and Models}
\label{sec:models}

We study three recent end-to-end RE models on \textbf{CoNLL04} \cite{roth-yih-2004-linear}, \textbf{ACE05} \cite{Walker2005} and \textbf{SciERC} \cite{luan-etal-2018-multi}.
They rely on various pretrained LMs and for a fairer comparison, we use BERT \cite{devlin-etal-2019-bert} on ACE05 and CoNLL04 and SciBERT \cite{beltagy-etal-2019-scibert} on SciERC\footnotemark[1] .

\footnotetext[1]{More implementation details in \autoref{app:implementation}}

\paragraph{PURE} \cite{Zhong2021AExtraction}   follows the pipeline approach. 
The NER model is a classical span-based model \cite{sohrab-miwa-2018-deep}. Special tokens corresponding to each predicted entity span are added and used as representation for Relation Classification. For a fairer comparison with other models, we study the approximation model that only requires one pass in each encoder and limits to sentence-level prediction. However, it still requires finetuning and storing two pretrained LMs instead of a single one for the following models.

\paragraph{SpERT} \cite{Eberts2020Span-basedPre-training}  uses a similar span-based NER module. 
RE is performed based on the filtered representations of candidate arguments as well as a max-pooled representation of their middle context. While Entity Filtering is close to the pipeline approach, the NER and RE modules share a common entity representation and are trained jointly. 
We also study the ablation of the max-pooled context representation that we denote \textbf{Ent-SpERT}.

\paragraph{Two are better than one (TABTO)} \cite{wang-lu-2020-two}  intertwines a sequence encoder and a table encoder in a Table Filling approach \cite{miwa-sasaki-2014-modeling}.
Contrary to previous models the pretrained LM is frozen and both the final hidden states and attention weights are used by the encoders. The prediction is finally performed by a Multi-Dimensional RNN (MD-RNN). Because it is not based on span-level predictions, this model cannot detect nested entities, e.g. on SciERC.

    \begin{table*}[t]
    \centering
    \small
    
    \begin{tabularx}{\textwidth}{@{}l*{3}{Y}r*{4}{Y}r*{4}{Y}c@{}}
    
    \toprule
    \multicolumn{1}{c}{\multirow{2}{*}{$\mu$ F1}}
    
     & \multicolumn{3}{c}{NER} &  & \multicolumn{4}{c}{RE Boundaries} & & \multicolumn{4}{c}{RE Strict} \\
    \cline{2-4} \cline{6-9} \cline{11-14} 
    & Seen & Unseen & All & & Exact & Partial & New & All &  &  Exact & Partial & New & All  \\
    
    \midrule
    & \multicolumn{14}{c}{\textbf{ACE05}} \\
    \cline{2-4} \cline{6-9} \cline{11-14} 
    \textit{proportion} &\textit{ 82\%} & \textit{18\%} & & & \textit{23\%} & \textit{63\%} & \textit{14\%} & & & \textit{23\%} & \textit{63\%}& \textit{14\%}\\
    \midrule
    heuristic & 59.2$_{\stddev{\phantom{0.0}}}$ & - & 55.1$_{\stddev{\phantom{0.0}}}$ &  & 37.9$_{\stddev{\phantom{0.0}}}$ & - & - & 23.0$_{\stddev{\phantom{0.0}}}$ &  & 34.3$_{\stddev{\phantom{0.0}}}$ & - & - & 20.8$_{\stddev{\phantom{0.0}}}$ &  \\
    
    Ent-SpERT & 89.0$_{\stddev{0.1}}$ & 74.1$_{\stddev{1.0}}$ & 86.5$_{\stddev{0.2}}$ &  & 77.0$_{\stddev{1.1}}$ & 52.2$_{\stddev{1.1}}$ & 38.9$_{\stddev{1.0}}$ & 57.0$_{\stddev{0.8}}$ &  & 75.1$_{\stddev{1.2}}$ & 48.4$_{\stddev{1.0}}$ & 36.3$_{\stddev{2.0}}$ & 53.9$_{\stddev{0.8}}$ &  \\
    
    SpERT & 89.4$_{\stddev{0.2}}$ & 74.2$_{\stddev{0.8}}$ & 86.8$_{\stddev{0.2}}$ &  & 84.8$_{\stddev{0.8}}$ & 59.6$_{\stddev{0.7}}$ & 42.3$_{\stddev{1.1}}$ & 64.0$_{\stddev{0.6}}$ &  & 82.6$_{\stddev{0.8}}$ & 55.6$_{\stddev{0.7}}$ & 38.4$_{\stddev{1.1}}$ & 60.6$_{\stddev{0.5}}$ &  \\
    
    
    TABTO & 89.7$_{\stddev{0.1}}$ & 77.4$_{\stddev{0.8}}$ & 87.5$_{\stddev{0.2}}$ &  & 85.9$_{\stddev{0.9}}$ & 62.6$_{\stddev{1.8}}$ & 44.6$_{\stddev{2.9}}$ & \textbf{66.4}$_{\stddev{1.3}}$ &  & 81.6$_{\stddev{1.5}}$ & 58.1$_{\stddev{1.6}}$ & 38.5$_{\stddev{3.1}}$ & \textbf{61.7}$_{\stddev{1.1}}$ &  \\
    
    

    PURE & 90.5$_{\stddev{0.2}}$ & 80.0$_{\stddev{0.3}}$ & \textbf{88.7}$_{\stddev{0.1}}$ &  & 86.0$_{\stddev{1.3}}$ & 60.5$_{\stddev{1.0}}$ & 47.1$_{\stddev{1.6}}$ & \textbf{65.1}$_{\stddev{0.7}}$ &  & 84.1$_{\stddev{1.1}}$ & 57.9$_{\stddev{1.3}}$ & 44.0$_{\stddev{2.0}}$ & \textbf{62.6}$_{\stddev{0.9}}$ &  \\
    
    \midrule
     & \multicolumn{14}{c}{\textbf{CoNLL04}} \\
    \cline{2-4} \cline{6-9} \cline{11-14} 
    \textit{proportion} &\textit{ 50\%} & \textit{50\%} & & & \textit{23\%} & \textit{34\%} & \textit{43\%} & & & \textit{23\%} & \textit{34\%}& \textit{43\%} \\
    \midrule
    heuristic & 86.0$_{\stddev{\phantom{0.0}}}$ & - & 59.7$_{\stddev{\phantom{0.0}}}$ &  & 90.9$_{\stddev{\phantom{0.0}}}$ & - & - & 35.5$_{\stddev{\phantom{0.0}}}$ &  & 90.9$_{\stddev{\phantom{0.0}}}$ & - & - & 35.5$_{\stddev{\phantom{0.0}}}$ &  \\
     
    Ent-SpERT & 95.9$_{\stddev{0.3}}$ & 81.9$_{\stddev{0.2}}$ & \textbf{88.9}$_{\stddev{0.2}}$ &  & 92.3$_{\stddev{1.4}}$ & 60.8$_{\stddev{1.4}}$ & 54.6$_{\stddev{1.3}}$ & 64.8$_{\stddev{0.9}}$ &  & 92.3$_{\stddev{1.4}}$ & 60.8$_{\stddev{1.4}}$ & 54.2$_{\stddev{1.2}}$ & 64.7$_{\stddev{0.8}}$ &  \\
    
    SpERT & 95.4$_{\stddev{0.4}}$ & 81.2$_{\stddev{0.4}}$ & 88.3$_{\stddev{0.2}}$ &  & 91.4$_{\stddev{0.6}}$ & 67.0$_{\stddev{1.1}}$ & 59.0$_{\stddev{1.4}}$ & 69.3$_{\stddev{1.2}}$ &  & 91.4$_{\stddev{0.6}}$ & 66.9$_{\stddev{1.1}}$ & 58.5$_{\stddev{1.4}}$ & 69.0$_{\stddev{1.2}}$ &  \\
   
    TABTO & 95.4$_{\stddev{0.4}}$ & 83.1$_{\stddev{0.7}}$ & \textbf{89.2}$_{\stddev{0.5}}$ &  & 92.6$_{\stddev{1.5}}$ & 72.6$_{\stddev{2.1}}$ & 64.8$_{\stddev{1.0}}$ & \textbf{74.0}$_{\stddev{1.4}}$ &  & 92.6$_{\stddev{1.5}}$ & 72.1$_{\stddev{1.8}}$ & 64.7$_{\stddev{1.1}}$ & \textbf{73.8}$_{\stddev{1.2}}$ &  \\
    
    PURE & 95.0$_{\stddev{0.2}}$ & 81.8$_{\stddev{0.2}}$ & 88.4$_{\stddev{0.2}}$ &  & 90.1$_{\stddev{1.3}}$ & 66.6$_{\stddev{1.0}}$ & 58.6$_{\stddev{1.5}}$ & 68.3$_{\stddev{1.0}}$ &  & 89.9$_{\stddev{1.4}}$ & 66.6$_{\stddev{1.0}}$ & 58.5$_{\stddev{1.5}}$ & 68.2$_{\stddev{0.9}}$ &  \\
    
    \midrule
    & \multicolumn{14}{c}{\textbf{SciERC}} \\
    \cline{2-4} \cline{6-9} \cline{11-14} 
    \textit{proportion} &\textit{ 23\%} & \textit{77\%} & & & \textit{<1\%} & \textit{30\%} & \textit{69\%} & & & \textit{<1\%} & \textit{30\%}& \textit{69\%} \\
    \midrule
    
    heuristic & 31.3$_{\stddev{\phantom{0.0}}}$ & - & 20.1$_{\stddev{\phantom{0.0}}}$ & 
    & - & - & - & 0.7$_{\stddev{\phantom{0.0}}}$ &  
    & - & - & - & 0.7$_{\stddev{\phantom{0.0}}}$ \\
    
    Ent-SpERT & 77.6$_{\stddev{1.0}}$ & 64.0$_{\stddev{0.6}}$ & \textbf{67.3}$_{\stddev{0.6}}$ &  
    & - & 48.1$_{\stddev{0.7}}$ & 41.9$_{\stddev{0.6}}$ & 43.8$_{\stddev{0.5}}$ &  
    & - & 38.1$_{\stddev{1.9}}$ & 29.4$_{\stddev{1.1}}$ & 32.1$_{\stddev{1.2}}$ & \\
    
    SpERT & 78.5$_{\stddev{0.5}}$ & 64.2$_{\stddev{0.4}}$ & \textbf{67.6}$_{\stddev{0.3}}$ &  
    & - & 53.1$_{\stddev{1.2}}$ & 46.0$_{\stddev{1.0}}$ & \textbf{48.2}$_{\stddev{1.1}}$ &  
    & - & 43.0$_{\stddev{1.6}}$ & 33.2$_{\stddev{1.1}}$ & \textbf{36.2}$_{\stddev{1.0}}$ & \\
    
    PURE & 78.0$_{\stddev{0.5}}$ & 63.8$_{\stddev{0.6}}$ & \textbf{67.2}$_{\stddev{0.4}}$& & - & 54.0$_{\stddev{0.7}}$ & 44.8$_{\stddev{0.4}}$ & \textbf{47.6}$_{\stddev{0.3}}$ & 
    & - & 42.2$_{\stddev{0.7}}$ & 32.6$_{\stddev{0.7}}$ & \textbf{35.6}$_{\stddev{0.6}}$ & \\
    
    \bottomrule
    
    \end{tabularx}
    \caption{Test NER and RE F1 Scores separated by lexical overlap with the training set. Exact Match RE scores are not reported on SciERC where the support is composed of only 5 exactly seen relation instances. Average and standard deviations on five runs. 
    }
    \label{table:overlap}
    \end{table*}

\section{Partitioning by Lexical Overlap}

Following \cite{Augenstein2017GeneralisationAnalysis, Taille2020ContextualizedGeneralization}, we partition the entity mentions in the test set based on lexical overlap with the training set.
We distinguish \textit{Seen} and \textit{Unseen} mentions and also extend this partition to relations.
We denote a relation as an \textit{Exact Match} if the same (head, predicate, tail) triple appears in the train set; as a \textit{Partial Match} if one of its arguments appears in the same position in a training relation of same type; and as \textit{New} otherwise.

We implement a naive \textbf{Retention Heuristic} that tags an entity mention or a relation exactly present in the training set with its majority label.
We report micro-averaged Precision, Recall and F1 scores for both NER and RE in \autoref{table:overlap}.

An entity mention is considered correct if both its boundaries and type have been correctly predicted.
For RE, we report scores in the \textbf{Boundaries} and \textbf{Strict} settings \cite{bekoulis-etal-2018-adversarial, Taille2020LetsExtraction}.
In the Boundaries setting, a relation is correct if its type is correct and the boundaries of its arguments are correct, without considering the detection of their types.
The Strict setting adds the requirement that the entity type of both argument is correct.

\subsection{Dataset Specificities}
We first observe very different statistics of Mention and Relation Lexical Overlap in the three datasets, which can be explained by the singularities of their entities and relations.
In CoNLL04, mentions are mainly Named Entities denoted with proper names while in ACE05 the surface forms are very often common names or even pronouns, which explains the occurrence of training entity mentions such as "it", "which", "people" in test examples.
This also leads to a weaker entity label consistency \cite{fu-etal-2020-interpretable}: "it" is labeled with every possible entity type and appears mostly unlabeled  whereas a mention such as "President Kennedy" is always labeled as a person in CoNLL04.
Similarly, mentions in SciERC are common names which can be tagged with different labels and they can also be nested.
Both the poor label consistency as well as the nested nature of entities hurt the performance of the retention heuristic.

For RE, while SciERC has almost no exact overlap between test and train relations, ACE05 and CoNLL04 have similar levels of exact match.
The larger proportion of partial match in ACE05 is explained by the pronouns that are more likely to co-occur in several instances.
The difference in performance of the heuristic is also explained by a poor relation label consistency.

\subsection{Lexical Overlap Bias}
As expected, this first evaluation setting enables to expose an important lexical overlap bias, already discussed in NER, in end-to-end Relation Extraction.
On every dataset and for every model micro F1 scores are the highest for Exact Match relations, then Partial Match and finally totally unseen relations.
This is a first confirmation that retention plays an important role in the measured overall performance of end-to-end RE models.

\subsection{Model Comparisons}
While we cannot evaluate TABTO on SciERC because it is unfit for extraction of nested entities, we can notice different hierarchies of models on every dataset suggesting that there is no one-size-fits-all best model, at least in current evaluation settings.

The most obvious comparison is between SpERT and Ent-SpERT where the explicit representation of context is ablated.
This results in a loss of performance on the RE part and especially on partially matching or new relations for which the entity representations pairs have not been seen.
Ent-SpERT is particularly effective on Exact Matches on CoNLL04, suggesting its retention capability.

Other comparisons are more difficult, given the numerous variations between the very structure of each model as well as training procedures.
However, the PURE pipeline setting seems to only be more effective on ACE05 where its NER performance is significantly better, probably because learning a separate NER and RE encoder enables to learn and capture more specific information for each distinctive task.
Even then, TABTO yields better Boundaries performance only penalized on the Strict setting by entity types confusions.
On the contrary, on CoNLL04, TABTO significantly outperforms its counterparts, especially on unseen relations.
This indicates that it proposes a more effective incorporation of contextual information in this case where relation and argument types are mapped bijectively.

On SciERC, performance of all models is already compromised at the NER level before the RE step, which makes further distinction between model performance even more difficult.

    \begin{table*}[h]
    
    
    \begin{tabularx}{\textwidth}{@{}lXX@{}}
    
    \toprule
    & Sentence & Ground Truth Relation \\
    \midrule
    Original & \textbf{John Wilkes Booth} , who assassinated \textbf{President Lincoln} , was an actor . & (John Wilkes Booth, Kill, President Lincoln) \\
    \midrule
    Swapped & \textbf{President Lincoln} , who assassinated \textbf{John Wilkes Booth} , was an actor . & (President Lincoln, Kill, John Wilkes Booth) \\
    \bottomrule
    
    \end{tabularx}
    \caption{Example of Swapped sentence. The Triple (John Wilkes Booth, Kill, President Lincoln) is present in the training set and the retention behaviours lead models to extract this triple when probed with the swapped sentence expressing the reverse relation.}
    
    \label{table:ex_swap}
    \end{table*}

    \begin{table}[t]
    \centering
    \small
    
    \begin{tabularx}{\columnwidth}{@{}l*{2}{Y}r*{2}{Y}r*{2}{Y}c@{}}
    
    \toprule
     
    \multicolumn{1}{c}{\multirow{2}{*}{F1}} & \multicolumn{2}{c}{NER $\uparrow$} &  & \multicolumn{2}{c}{RE $\uparrow$} & & \multicolumn{2}{c}{revRE $\downarrow$} \\
      
     \cline{2-3} \cline{5-6} \cline{8-9}
     
     & O & S & & O & S & & O & S &\\
    
    \midrule
    \multicolumn{10}{c}{Kill} \\
    \midrule
     
    Ent-SpERT & 91.6 & 91.7 &  & 85.1 & 35.4 &  & - & 58.5 & \\

    SpERT & 91.4 & \textbf{92.6} &  & 86.2 & 35.0 &  & - & 57.8 & \\
    
    TABTO & \textbf{92.0} & \textbf{92.8} &  & \textbf{89.6} & 27.6 &  & - & 59.5 & \\
    
    PURE & 90.5 & 90.7 &  & 84.1 & \textbf{52.3} &  & - & \textbf{14.3} & \\

    \midrule
    \multicolumn{10}{c}{Located in} \\
    \midrule
    
    Ent-SpERT & \textbf{90.0} & 87.0 &  & 78.3 & 30.3 &  & - & 24.8 & \\
    
    SpERT & 88.6 & 87.7 &  & 75.0 & 24.9 &  & - & 33.5 & \\
    
    TABTO & \textbf{90.1} & \textbf{88.9} &  & \textbf{85.3} & 36.1 &  & - & 34.9 & \\
    
    PURE & 89.0 & 83.7 &  & 81.2 & \textbf{59.3} &  & - & \textbf{5.1} & \\
    \bottomrule
    
    \end{tabularx}
    \caption{Performance on CoNLL04 test set containing exactly one relation of the corresponding type in its original form (O) and where the relation head and tail are swapped (S). NER F1 score is micro-averaged while strict RE score only takes these relations into account. The revRE score corresponds to unwanted extraction of the reverse relation, symptomatic of the retention effect in the swapped setting.}
    \label{table:swap_rev}
    \end{table}

\section{Swapping Relation Heads and Tails}
A second experiment to validate that retention is used as a heuristic in models' predictions is to modify their input sentences in a controlled manner similarly to what is proposed in \cite{ribeiro-etal-2020-beyond}.
We propose a very focused experiment that consists in selecting asymmetric relations that occur between entities of same type and swap the head with the tail in the input.
If the model predicts the original triple, then it over relies on the retention heuristic, whereas finding the swapped triple is an evidence of broader context incorporation. We show an example in \autoref{table:ex_swap}.

Because of the requirements of this experiment, we have to limit to two relations in CoNLL04: ``Kill'' between people and ``Located in'' between locations.
Indeed, CoNLL04 is the only dataset with a bijective mapping between the type of a relation and the types of its arguments and the consistent proper nouns mentions makes the swaps mostly grammatically correct. 
For each relation type, we only consider sentences with exactly one instance of corresponding relation and swap its arguments. 
We only consider this relation in the RE scores reported in \autoref{table:swap_rev}.
We use the strict RE score as well as \textbf{revRE} which measures the extraction of the reverse relation, not expressed in the sentence.

For each relation, the hierarchy of models corresponds to the overall CoNLL04. 
Swapping arguments has a limited effect on NER, mostly for the "Located in" relation.
However, it leads to a drop in RE for every model and the revRE score indicates that SpERT and TABTO predict the reverse relation more often than the newly expressed one.
This is another proof of the retention heuristic of end-to-end models, although it might also be attributed to the language model to the language model. In particular for the ''Located in`` relation, swapped heads and tails are not exactly equivalent since the former are mainly cities and the latter countries.

On the contrary, the PURE model is less prone to information retention, as shown by its revRE scores significantly smaller than the standard RE scores on swapped sentences. Hence, it outperforms SpERT and TABTO on swapped sentences despite being the least effective on the original dataset.The important discrepancy in results can be explained by the different types of representations used by these models. The pipeline approach allows the use of argument type representations in the Relation Classifier whereas most end-to-end models use lexical features in a shared entity representation used for both NER and RE.

These conclusions from quantitative results are validated qualitatively.
We can observe that the four predominant patterns are intuitive behaviours on sentences with swapped relations: retention of the incorrect original triple, prediction of the correct swapped triple and prediction of none or both triples.
We report some examples in \autoref{table:qualitative_kill} and \autoref{table:qualitative_loc} in the Appendix.

\section{Related Work}
\label{sec:related_work}

Several works on generalization of NER models mention lexical overlap with the training as a key indicator of performance.
\citet{Augenstein2017GeneralisationAnalysis} separate mentions in the test set as seen and unseen during training and measure out-of-domain generalization in an extensive study of two CRF based models and SENNA combining a Convolutional Neural Network with a CRF \citep{Collobert2011NaturalScratch}.
\citet{Taille2020ContextualizedGeneralization} compare the effect of introducing contextual embeddings in the classical BiLSTM-CRF architecture in a similar setting and show that they help close the performance gap on unseen mentions and domains. \citet{arora-etal-2020-contextual, Fu2020RethinkingStudy, fu-etal-2020-interpretable} study the influence of several properties such as lexical overlap, label consistency and entity length on state-of-the-art models performance. They model these properties as continuous scores associated to each mention and bucketized for evaluation.
Lexical overlap has also been mentioned in Coreference Resolution \cite{moosavi-strube-2017-lexical} where coreferent mentions tend to co-occur in the test and train sets.
In this line of works, the impact of lexical overlap is measured either by separating performance depending on the property of mentions (seen or unseen) or with out-of-domain evaluation with a test set from a different dataset with lower lexical overlap with the train set.

Another recently proposed method for fine-grained evaluation of NLP models beyond a single benchmark score is to modify the test sentences in a controlled manner.
\citet{mccoy-etal-2019-right} expose lexical overlap as a shallow heuristic adopted by state-of-the-art Natural Language Inference models, especially by swapping subject and object of verbs in the hypothesis of some examples where the premise entails the hypothesis. While such a modification changes the label of these examples to non-entailment, all models tested show a spectacular drop of accuracy on these models.
\citet{ribeiro-etal-2020-beyond} propose a broader set of test set modifications to individually test robustness of NLP models to several patterns such as the introduction of negation, swapping words with synonyms, changing tense and much more.

In pipeline RE where ground truth candidate arguments are given, models often use intermediate representations based on entity types that reduce lexical overlap issues. However, \citet{rosenman-etal-2020-exposing} show that they still tend to adopt shallow heuristics based on the type of the arguments and the presence of triggers indicative of the presence of a relation. They propose hard cases with several mentions of same types for which Relation Classifiers struggle connecting the correct pair. Concurrently, \citet{peng-etal-2020-learning} confirm that RE benchmarks present shallow cues such as the type of the candidate arguments that can be used alone to infer the relation. 

We propose to extend previous work on NER and RE to the more realistic end-to-end RE setting with two of the previously described approaches: 1) separating performance by lexical overlap of mentions or argument pairs and 2) modifying some CoNLL04 test examples by swapping relations heads and tails.

\section{Conclusion}

In this paper, we study three state-of-the-art end-to-end Relation Extraction models in order to highlight their tendency to retain seen relations.
We confirm that retention of seen mentions and relations play an important role in overall RE performance and can explain the relatively higher scores on CoNLL04 and ACE05 compared to SciERC.
Furthermore, our experiment on swapping relation heads and tails tends to show that the intermediate manipulation of type representations instead of lexical features enabled in the pipeline PURE model makes it less prone to over-rely on retention.

While the limited extend of our swapping experiment is an obvious limitation of this work, it shows limitations of both current benchmarks and models. It is an encouragement to propose new benchmarks that might be easily modified by design to probe such lexical overlap heuristics.
Contextual information could for example be contained in templates of that would be filled with different (head, tail) pairs either seen or unseen during training.

Furthermore, pretrained Language Models can already capture relational information between phrases \citep{petroni-etal-2019-language} and further experiments could help distinguish their role in the retention behaviour of RE models.

\section*{Acknowledgments}
We thank the anonymous reviewers.
This work was performed while Bruno Taillé was employed by BNP Paribas and supported by the French \mbox{Ministry} for Research (CIFRE convention 2018/0327).
We also thank the H2020 project AI4EU (825619) which partially supports Patrick Gallinari.
\clearpage

\bibliography{biblio/anthology, biblio/references, biblio/custom}
\bibliographystyle{emnlp_template/acl_natbib}

\clearpage
\appendix

\section{Implementation Details}
\label{app:implementation}

For every model, we use the original code associated with the papers with the default best performing hyperparameters unless stated otherwise.
We run 5 runs on a single NVIDIA 2080Ti GPU for each of them on each dataset. 
For CoNLL04 and ACE05, we train each model with both the cased and uncased versions of BERT$_{BASE}$ and only keep the best performing setting. 

\paragraph{PURE} \cite{Zhong2021AExtraction} \footnotemark[1] 
We use the approximation model and limit use a \textit{context window} of 0 to only use the current sentence for prediction and be able to compare with other models.
For ACE05, we use the standard\textit{ bert-base-uncased} LM but use the \textit{bert-base-cased} version on CoNLL04 which results in a significant +2.4 absolute improvement in RE Strict micro F1 score.

\paragraph{SpERT} \cite{Eberts2020Span-basedPre-training} \footnotemark[2] We use the original implementation as is with \textit{bert-base-cased} for both ACE05 and CoNLL04 since the uncased version is not beneficial, even on ACE05 where there are fewer proper nouns.
For the Ent-SpERT ablation, we simply remove the max-pooled context representation from the final concatenation in the RE module. 
This modifies the RE classifier's input dimension from the original 2354 to 1586.

\paragraph{Two are better than one (TABTO)} \cite{wang-lu-2020-two} \footnotemark[3] We use the original implementation with \textit{bert-base-uncased} for both ACE05 and CoNLL04 since the cased version is not beneficial on CoNLL04.

\footnotetext[1]{\href{https://github.com/princeton-nlp/PURE}{github.com/princeton-nlp/PURE}}
\footnotetext[2]{\href{https://github.com/lavis-nlp/spert}{github.com/lavis-nlp/spert}}
\footnotetext[3]{\href{https://github.com/LorrinWWW/two-are-better-than-one}{github.com/LorrinWWW/two-are-better-than-one}}

\section{Datasets Statistics}

We present general datasets statistics in \autoref{table:data_stats}.

    \begin{table}[h]
    \centering
    
    \begin{tabularx}{\columnwidth}{@{}lRRR@{}}
    
    \toprule
   
    \textbf{ACE05} & Train & Dev & Test \\
  
    \midrule
    
    Sentences & 10,051 & 2,424 & 2,050 \\
    Mentions & 26,473 & 6,338 & 5,476 \\
    Relations & 4,788 & 1,131 & 1,151 \\
    
    \midrule
    \textbf{CoNLL04} & Train & Dev & Test \\
    \midrule
    
    Sentences & 922 & 231 & 288 \\
    Mentions & 3,377 & 893 & 1,079 \\
    Relations & 1,283 & 343 & 422 \\
    
    \midrule
    \textbf{SciERC} & Train & Dev & Test \\
    \midrule
    
    Sentences & 1,861 & 275 & 551 \\
    Mentions &  5,598 & 811 & 1,685 \\
    Relations & 3,219 & 455 & 974 \\
    
    \bottomrule
    
    \end{tabularx}
    \caption{Datasets Statistics}
    \label{table:data_stats}
    \end{table}

We also compute average values of some entity and relation attributes inspired by \cite{fu-etal-2020-interpretable} and reported in \autoref{table:data_consistency}.

We report two of their entity attributes: \textbf{entity length} in number of tokens (\textbf{eLen}) and \textbf{entity label consistency (eCon)}.
Given a test entity mention, its label consistency is the number of occurrences in the training set with the same type divided by its total number of occurrences.
It is zero for unseen mentions.
Because eCon reflects both the ambiguity of labels for seen entities and the proportion of unseen entities, we propose to introduce the \textbf{eCon*} score that only averages label consistency of seen mentions and \textbf{eLex}, the proportion of entities with lexical overlap with the train set.

We introduce similar scores for relations. \textbf{Relation label consistency (rCon)} extends label consistency for triples. \textbf{Argument types label constitency  (aCon)} considers the labels of every pair of mentions of corresponding types in the training set. Because pairs of types are all seen during training we do not decompose aCon into aCon* and aLex. \textbf{Argument length  (aLen)} is the sum of the lengths of the head and tail mentions. \textbf{Argument distance (aDist)} is the number of tokens between the head and the tail of a relation.

We present a more complete report of overall Precision, Recall and F1 scores that can be interpreted in light of these statistics in \autoref{table:overlap_prf}.

    \begin{table*}[t]
    
    \begin{tabularx}{\textwidth}{@{}l*{4}{Y}r*{8}{Y}@{}}
    
    \toprule
   
           & \multicolumn{4}{c}{Entities} & & \multicolumn{8}{c}{Relations} \\
            
            \cline{2-5} \cline{7-12}
            & eCon & eCon* & eLex & eLen & 
            & rCon & rCon* & rLex & aCon & aLen & aDist \\
           
    \midrule
    
    \textbf{ACE05}   & 65\% & 78\% & 82\% & 1.1 &
            & 15\% & 62\% & 23\%  & 7.1\% & 2.3 & 2.8 \\
    
    \textbf{CoNLL04} & 49\% & 98\% & 50\% & 1.5 &
            & 21\% & 91\% & 23\% & 29\% & 3.8 & 5.8 \\
            
    \textbf{SciERC}  & 17\% & 74\% & 23\% & 1.6 &
            & 0.4\% & 74\% & 0.5\% & 13\% & 4.7 & 5.3\\
            
    \bottomrule
    
    \end{tabularx}
    \caption{Average of some entity and relation attributes in the test set.}
    \label{table:data_consistency}
    \end{table*}

    \begin{table*}[h]
    \centering
    \small
    
    \begin{tabularx}{\textwidth}{@{}l*{3}{Y}r*{3}{Y}r*{3}{Y}c@{}}
    
    \toprule
    \multicolumn{1}{c}{\multirow{2}{*}{$\mu$ F1}}
    
     & \multicolumn{3}{c}{NER} &  & \multicolumn{3}{c}{RE Boundaries} & & \multicolumn{3}{c}{RE Strict} \\
    \cline{2-4} \cline{6-8} \cline{10-12} 
    & P & R & F1 & & P & R & F1 & & P & R & F1 &  \\
    
    \midrule
    & \multicolumn{12}{c}{\textbf{ACE05}} \\
    \midrule
    
    heuristic & 44.7$_{\stddev{\phantom{0.0}}}$ & 71.9$_{\stddev{\phantom{0.0}}}$ & 55.1$_{\stddev{\phantom{0.0}}}$ &  & 23.6$_{\stddev{\phantom{0.0}}}$ & 22.3$_{\stddev{\phantom{0.0}}}$ & 23.0$_{\stddev{\phantom{0.0}}}$ &  & 21.4$_{\stddev{\phantom{0.0}}}$ & 20.2$_{\stddev{\phantom{0.0}}}$ & 20.8$_{\stddev{\phantom{0.0}}}$ & \\

    Ent-SpERT & 86.7$_{\stddev{0.3}}$ & 86.3$_{\stddev{0.3}}$ & 86.5$_{\stddev{0.2}}$ &  & 56.7$_{\stddev{1.0}}$ & 57.4$_{\stddev{0.7}}$ & 57.0$_{\stddev{0.8}}$ &  & 53.5$_{\stddev{1.0}}$ & 54.2$_{\stddev{0.8}}$ & 53.9$_{\stddev{0.8}}$ & \\
    
    SpERT & 87.2$_{\stddev{0.2}}$ & 86.5$_{\stddev{0.3}}$ & 86.8$_{\stddev{0.2}}$ &  & 68.1$_{\stddev{1.1}}$ & 60.5$_{\stddev{0.5}}$ & 64.0$_{\stddev{0.6}}$ &  & 64.4$_{\stddev{1.1}}$ & 57.2$_{\stddev{0.4}}$ & 60.6$_{\stddev{0.5}}$ & \\
    
    TABTO & 86.7$_{\stddev{0.3}}$ & 88.3$_{\stddev{0.6}}$ & 87.5$_{\stddev{0.2}}$ &  & 71.0$_{\stddev{2.7}}$ & 62.5$_{\stddev{2.5}}$ & 66.4$_{\stddev{1.3}}$ &  & 66.1$_{\stddev{2.6}}$ & 58.1$_{\stddev{2.1}}$ & 61.8$_{\stddev{1.1}}$ & \\
    
    PURE & 88.8$_{\stddev{0.3}}$ & 88.6$_{\stddev{0.1}}$ & 88.7$_{\stddev{0.1}}$ &  & 67.4$_{\stddev{0.8}}$ & 63.0$_{\stddev{0.8}}$ & 65.1$_{\stddev{0.7}}$ &  & 64.8$_{\stddev{1.0}}$ & 60.5$_{\stddev{1.0}}$ & 62.6$_{\stddev{0.9}}$ & \\
 
    \midrule
     & \multicolumn{12}{c}{\textbf{CoNLL04}} \\
    \midrule
    
    heuristic & 75.9$_{\stddev{\phantom{0.0}}}$ & 49.2$_{\stddev{\phantom{0.0}}}$ & 59.7$_{\stddev{\phantom{0.0}}}$ &  & 84.1$_{\stddev{\phantom{0.0}}}$ & 22.5$_{\stddev{\phantom{0.0}}}$ & 35.5$_{\stddev{\phantom{0.0}}}$ &  & 84.1$_{\stddev{\phantom{0.0}}}$ & 22.5$_{\stddev{\phantom{0.0}}}$ & 35.5$_{\stddev{\phantom{0.0}}}$ & \\

Ent-SpERT & 88.4$_{\stddev{0.6}}$ & 89.3$_{\stddev{0.7}}$ & 88.9$_{\stddev{0.2}}$ &  & 59.3$_{\stddev{0.5}}$ & 71.3$_{\stddev{1.5}}$ & 64.8$_{\stddev{0.9}}$ &  & 59.2$_{\stddev{0.5}}$ & 71.2$_{\stddev{1.5}}$ & 64.7$_{\stddev{0.8}}$ & \\

SpERT & 87.9$_{\stddev{0.6}}$ & 88.7$_{\stddev{0.3}}$ & 88.3$_{\stddev{0.2}}$ &  & 69.7$_{\stddev{2.3}}$ & 69.0$_{\stddev{0.5}}$ & 69.3$_{\stddev{1.2}}$ &  & 69.4$_{\stddev{2.3}}$ & 68.7$_{\stddev{0.6}}$ & 69.0$_{\stddev{1.2}}$ & \\

TABTO & 89.0$_{\stddev{0.7}}$ & 89.3$_{\stddev{0.3}}$ & 89.2$_{\stddev{0.5}}$ &  & 75.6$_{\stddev{3.2}}$ & 72.6$_{\stddev{1.9}}$ & 74.0$_{\stddev{1.4}}$ &  & 75.4$_{\stddev{3.1}}$ & 72.4$_{\stddev{1.8}}$ & 73.8$_{\stddev{1.2}}$ & \\

PURE & 88.3$_{\stddev{0.4}}$ & 88.5$_{\stddev{0.5}}$ & 88.4$_{\stddev{0.2}}$ &  & 68.6$_{\stddev{2.0}}$ & 68.2$_{\stddev{1.6}}$ & 68.3$_{\stddev{1.0}}$ &  & 68.5$_{\stddev{2.0}}$ & 68.1$_{\stddev{1.5}}$ & 68.2$_{\stddev{0.9}}$ & \\

    \midrule
    & \multicolumn{12}{c}{\textbf{SciERC}} \\
   \midrule

    heuristic & 18.8$_{\stddev{\phantom{0.0}}}$ & 21.5$_{\stddev{\phantom{0.0}}}$ & 20.1$_{\stddev{\phantom{0.0}}}$ &  & 3.5$_{\stddev{\phantom{0.0}}}$ & 0.4$_{\stddev{\phantom{0.0}}}$ & 0.7$_{\stddev{\phantom{0.0}}}$ &  & 3.5$_{\stddev{\phantom{0.0}}}$ & 0.4$_{\stddev{\phantom{0.0}}}$ & 0.7$_{\stddev{\phantom{0.0}}}$ & \\

    Ent-SpERT & 68.0$_{\stddev{0.3}}$ & 66.6$_{\stddev{0.9}}$ & 67.3$_{\stddev{0.6}}$ &  & 44.8$_{\stddev{0.7}}$ & 42.9$_{\stddev{1.0}}$ & 43.8$_{\stddev{0.5}}$ &  & 32.9$_{\stddev{0.9}}$ & 31.5$_{\stddev{1.5}}$ & 32.1$_{\stddev{1.2}}$ & \\
    
    SpERT & 67.6$_{\stddev{0.5}}$ & 67.6$_{\stddev{0.2}}$ & 67.6$_{\stddev{0.3}}$ &  & 49.3$_{\stddev{1.4}}$ & 47.2$_{\stddev{1.3}}$ & 48.2$_{\stddev{1.1}}$ &  & 37.0$_{\stddev{1.3}}$ & 35.4$_{\stddev{1.0}}$ & 36.2$_{\stddev{1.0}}$ & \\
    
    PURE & 68.2$_{\stddev{0.6}}$ & 66.2$_{\stddev{0.9}}$ & 67.2$_{\stddev{0.4}}$ &  & 50.2$_{\stddev{0.9}}$ & 45.2$_{\stddev{1.0}}$ & 47.6$_{\stddev{0.3}}$ &  & 37.6$_{\stddev{1.2}}$ & 33.8$_{\stddev{0.7}}$ & 35.6$_{\stddev{0.6}}$ & \\

    \bottomrule
    
    \end{tabularx}
    \caption{Overall micro-averaged Test NER and Strict RE Precision, Recall and F1 scores.  Average and standard deviations on five runs.
    We can observe that the recall of the heuristic is correlated with the proportions of seen entities or triples (eLex or rLex).
    Its particularly high precision on CoNLL04 seems rather linked to the important label consistency of seen entities and relation (eCon* and rCon*).}
    \label{table:overlap_prf}
    \end{table*}

    \begin{table*}[h]
    \centering
    
    \begin{tabularx}{\textwidth}{@{}lXX@{}}
    
    \toprule
   
   Dataset  & Entity Types & Relation Types \\
  
    \midrule
   
    ACE05   & Facility, Geo-political Entity, Location, Person, Vehicle, Weapon 
            & Artifact, Gen-affiliation, Org-affiliation, Part-whole, Person-social, Physical \\
    
    \midrule
    CoNLL04 & Location, Organization, Other, Person 
            & Kill, Live in, Located in, Organization based in, Work for \\
            
    \midrule
    SciERC & Generic, Material, Method, Metric, Other Scientific Term, Task  
            & Compare*, Conjunction*, Evaluate for, Feature of, Hyponym of, Part of, Used for \\

    \bottomrule
    
    \end{tabularx}
    \caption{Entity and Relation Types of end-to-end RE datasets. SciERC presents two types of symmetric relations denoted with a *.}
    \label{table:data_types}
    \end{table*}

\clearpage

    \begin{table*}[h]
    \centering
    \small
    
    \begin{tabularx}{\textwidth}{@{}lll*{3}{Y}r*{3}{Y}r*{3}{Y}c@{}}
    
    \toprule
     
     & & & \multicolumn{3}{c}{NER $\uparrow$} &  & \multicolumn{3}{c}{RE Strict $\uparrow$} &  & \multicolumn{3}{c}{Reverse RE Strict $\downarrow$} \\
      
     \cline{4-6} \cline{8-10} \cline{12-14}
     & & & P & R & F1 & & P & R & F1 & & P & R & F \\
     \midrule
     
    \parbox[t]{2mm}{\multirow{8}{*}{\rotatebox[origin=c]{90}{Kill}}} 
    & \parbox[t]{2mm}{\multirow{4}{*}{\rotatebox[origin=c]{90}{Original}}}

    & Ent-SpERT & 91.7$_{\stddev{0.4}}$ & 91.5$_{\stddev{0.7}}$ & 91.6$_{\stddev{0.4}}$ &  & 82.9$_{\stddev{2.7}}$ & 87.6$_{\stddev{1.8}}$ & 85.1$_{\stddev{0.9}}$ &  & - & - & - & \\
    
    & & SpERT & 91.7$_{\stddev{2.1}}$ & 91.0$_{\stddev{1.0}}$ & 91.4$_{\stddev{1.2}}$ &  & 88.1$_{\stddev{3.1}}$ & 84.4$_{\stddev{1.4}}$ & 86.2$_{\stddev{1.4}}$ &  & - & - & - & \\

    & & TABTO & 91.8$_{\stddev{0.6}}$ & 92.2$_{\stddev{0.5}}$ & \textbf{92.0}$_{\stddev{0.4}}$ &  & 88.8$_{\stddev{1.6}}$ & 90.7$_{\stddev{3.3}}$ & \textbf{89.6}$_{\stddev{1.3}}$ &  & - & - & - & \\
    
    & & PURE & 91.5$_{\stddev{0.9}}$ & 89.6$_{\stddev{0.6}}$ & 90.5$_{\stddev{0.6}}$ &  & 87.2$_{\stddev{2.1}}$ & 81.3$_{\stddev{1.1}}$ & 84.1$_{\stddev{1.2}}$ &  & - & - & - & \\
    
    \cline{2-14}
    & \parbox[t]{2mm}{\multirow{4}{*}{\rotatebox[origin=c]{90}{Swap}}} 
    
    & Ent-SpERT & 91.3$_{\stddev{0.9}}$ & 92.1$_{\stddev{0.7}}$ & 91.7$_{\stddev{0.7}}$ &  & 31.8$_{\stddev{5.3}}$ & 40.0$_{\stddev{8.3}}$ & 35.4$_{\stddev{6.5}}$ &  & 52.8$_{\stddev{5.6}}$ & 65.8$_{\stddev{7.2}}$ & 58.5$_{\stddev{5.7}}$ & \\
    
    & & SpERT & 92.6$_{\stddev{1.8}}$ & 92.6$_{\stddev{0.8}}$ & \textbf{92.6}$_{\stddev{1.2}}$ &  & 33.0$_{\stddev{4.4}}$ & 37.3$_{\stddev{7.4}}$ & 35.0$_{\stddev{5.6}}$ &  & 54.8$_{\stddev{5.1}}$ & 61.3$_{\stddev{4.1}}$ & 57.8$_{\stddev{4.0}}$ & \\

    & & TABTO & 92.8$_{\stddev{0.8}}$ & 92.7$_{\stddev{0.9}}$ & \textbf{92.8}$_{\stddev{0.7}}$ &  & 26.8$_{\stddev{3.6}}$ & 28.4$_{\stddev{4.1}}$ & 27.6$_{\stddev{3.8}}$ &  & 57.8$_{\stddev{3.1}}$ & 61.3$_{\stddev{3.0}}$ & 59.5$_{\stddev{2.8}}$ & \\
    
    & & PURE & 92.0$_{\stddev{0.5}}$ & 89.5$_{\stddev{1.0}}$ & 90.7$_{\stddev{0.5}}$ &  & 65.2$_{\stddev{6.0}}$ & 44.0$_{\stddev{7.4}}$ & \textbf{52.3}$_{\stddev{6.5}}$ &  & 17.8$_{\stddev{2.3}}$ & 12.0$_{\stddev{2.3}}$ & \textbf{14.3}$_{\stddev{2.2}}$ & \\

    \midrule
    
    \parbox[t]{2mm}{\multirow{8}{*}{\rotatebox[origin=c]{90}{Located in}}} 
    & \parbox[t]{2mm}{\multirow{4}{*}{\rotatebox[origin=c]{90}{Original}}} 
    
    & Ent-SpERT & 90.1$_{\stddev{0.8}}$ & 89.8$_{\stddev{1.5}}$ & \textbf{90.0}$_{\stddev{0.7}}$ &  
                & 80.8$_{\stddev{3.7}}$ & 76.2$_{\stddev{3.2}}$ & 78.3$_{\stddev{2.4}}$ &  
                & - & - & - & \\
    
    & & SpERT   & 89.8$_{\stddev{1.2}}$ & 87.5$_{\stddev{1.5}}$ & 88.6$_{\stddev{1.1}}$ &  
                & 77.2$_{\stddev{2.8}}$ & 73.0$_{\stddev{3.0}}$ & 75.0$_{\stddev{2.0}}$ & 
                & - & - & - & \\

    & & TABTO   & 90.1$_{\stddev{1.3}}$ & 90.0$_{\stddev{1.8}}$ & \textbf{90.1}$_{\stddev{1.5}}$ &  
                & 93.0$_{\stddev{3.3}}$ & 78.9$_{\stddev{4.6}}$ & \textbf{85.3}$_{\stddev{3.9}}$ & 
                & - & - & - & \\

    & & PURE    & 88.6$_{\stddev{1.1}}$ & 89.4$_{\stddev{1.8}}$ & 89.0$_{\stddev{1.0}}$ &  
                & 89.3$_{\stddev{4.0}}$ & 74.6$_{\stddev{3.7}}$ & 81.2$_{\stddev{2.6}}$ & 
                & - & - & - & \\
    
    \cline{2-14}
    & \parbox[t]{2mm}{\multirow{4}{*}{\rotatebox[origin=c]{90}{Swap}}} 
    
    &  Ent-SpERT    & 86.7$_{\stddev{1.9}}$ & 87.4$_{\stddev{2.7}}$ & 87.0$_{\stddev{2.1}}$ &  
                    & 38.0$_{\stddev{8.5}}$ & 25.4$_{\stddev{2.8}}$ & 30.3$_{\stddev{4.6}}$ &  
                    & 30.2$_{\stddev{5.2}}$ & 21.1$_{\stddev{5.8}}$ & 24.8$_{\stddev{5.7}}$ & \\
    
    & & SpERT       & 87.3$_{\stddev{1.4}}$ & 88.0$_{\stddev{0.9}}$ & 87.7$_{\stddev{1.1}}$ & 
                    & 34.8$_{\stddev{14.8}}$ & 19.5$_{\stddev{6.7}}$ & 24.9$_{\stddev{9.2}}$ &  
                    & 45.6$_{\stddev{17.0}}$ & 26.5$_{\stddev{10.5}}$ & 33.5$_{\stddev{13.0}}$ & \\

    & & TABTO       & 89.0$_{\stddev{0.6}}$ & 88.8$_{\stddev{0.9}}$ & \textbf{88.9}$_{\stddev{0.8}}$ &
                    & 46.5$_{\stddev{6.6}}$ & 29.7$_{\stddev{5.7}}$ & 36.1$_{\stddev{5.8}}$ &  
                    & 45.2$_{\stddev{5.2}}$ & 28.6$_{\stddev{3.7}}$ & 34.9$_{\stddev{3.6}}$ & \\

    & & PURE        & 82.7$_{\stddev{0.8}}$ & 84.6$_{\stddev{0.8}}$ & 83.7$_{\stddev{0.5}}$ &  
                    & 74.9$_{\stddev{7.6}}$ & 49.7$_{\stddev{4.7}}$ & \textbf{59.3}$_{\stddev{3.0}}$ &  
                    & 6.5$_{\stddev{1.8}}$ & 4.3$_{\stddev{1.3}}$ & \textbf{5.1}$_{\stddev{1.5}}$ & \\
    
    \bottomrule
    
    \end{tabularx}
    \caption{Detailed results of the Swap Relation Experiment with Precision, Recall and F1 scores.}
    \label{table:swap_prf_rev}
    \end{table*}

\clearpage

    \begin{table*}[h]
    
    
    \begin{tabularx}{\textwidth}{@{}lYY@{}}
    
    \toprule
     \textbf{1} & \multicolumn{2}{C{0.8\textwidth}}{The Warren Commission determined that on Nov. 22 , 1963 , \textbf{A} fired a high-powered rifle at \textbf{B} 's motorcade from the sixth floor of what is now the Dallas County Administration Building , where he worked .} \\
    \midrule
      A, B &  Lee Harvey Oswald, Kennedy  & Kennedy, Lee Harvey Oswald\\
     \midrule
    Ent-SpERT & (A,B) & \textcolor{red}{(B,A)} \\
    SpERT & (A,B) & \textcolor{red}{(B,A)} \\
    TABTO & (A,B) & \textcolor{red}{(B,A)} \\
    PURE & (A,B) & \textcolor{red}{(B,A)} \\
    \midrule
    
    \midrule
     \textbf{2} & \multicolumn{2}{C{0.8\textwidth}}{Today 's Highlight in History : Twenty years ago , on June 6 , 1968 , at 1 : 44 a.m. local time , \textbf{B} died at Good Samaritan Hospital in Los Angeles , 25 -LCB- hours after he was shot at the Ambassador Hotel by \textbf{A} .} \\
    \midrule
     A, B &  Sirhan Bishara Sirhan, Sen. Robert F. Kennedy  &  Sen. Robert F. Kennedy, Sirhan Bishara Sirhan \\
     \midrule
     
    Ent-SpERT & (A,B) & \textcolor{red}{(B,A)} \\
    SpERT & (A,B) & \textcolor{red}{(B,A)} \\
    TABTO & (A,B) & \textcolor{red}{(B,A)} \\
    PURE & (A,B) & - \\
    \midrule
    
    \midrule
     \textbf{3} & \multicolumn{2}{C{0.8\textwidth}}{In 1968 , authorities announced the capture in London of \textbf{A} , suspected of the assassination of civil rights leader \textbf{B} .} \\
    \midrule
     A, B &  James Earl Ray, Dr. Martin Luther King Jr  & Dr. Martin Luther King Jr, James Earl Ray \\
     \midrule
     
    Ent-SpERT & (A,B) & (A,B) \textcolor{red}{(B,A)} \\
    SpERT & (A,B) & (A,B) \textcolor{red}{(B,A)} \\
    TABTO & (A,B) & (A,B) \\
    PURE & (A,B) & (A,B) \\
    \midrule
    
    \midrule
     \textbf{4} & \multicolumn{2}{C{0.8\textwidth}}{The Warren Commission determined that \textbf{A} fired at \textbf{B} from the sixth floor of what is now the Dallas County Administration Building .} \\
    \midrule
     A, B &  Oswald, Kennedy  & Kennedy, Oswald \\
     \midrule
     
    Ent-SpERT & (A,B) & - \\
    SpERT & (A,B) & (A,B) \textcolor{red}{(B,A)} \\
    TABTO & (A,B) & \textcolor{red}{(B,A)} \\
    PURE & (A,B) & (A,B) \\
    
    \bottomrule
    
    \end{tabularx}
    \caption{Some qualitative examples of models' predictions on original (left column) and swapped (right) CoNLL04 sentences for the ``Kill'' relation. Despite a perfect Relation Extraction in the original sentences for all models, swapping head and tails results in several types of errors mainly regarding the direction of the relation. Predictions of incorrect original triples are in red. These examples are obtained from models trained with the same seed ($s=0$).}
    
    \label{table:qualitative_kill}
    \end{table*}

    \begin{table*}[h]
    
    
    \begin{tabularx}{\textwidth}{@{}lYY@{}}
    
    \toprule
     \textbf{1} & \multicolumn{2}{C{0.8\textwidth}}{
Reagan recalled that on the 40th anniversary of the Normandy landings he read a letter from a young woman whose late father had fought at \textbf{A} , a \textbf{B} sector .} \\
    \midrule
      A, B &  Omaha Beach, Normandy  & Normandy, Omaha Beach\\
    \midrule
    Ent-SpERT & (A,B) & - \\
    SpERT & (A,B) & - \\
    TABTO & (A,B) & - \\
    PURE & (A,B) & (A,B) \\
    \midrule
    
    \midrule
     \textbf{2} & \multicolumn{2}{C{0.8\textwidth}}{\textbf{A} , \textbf{B} ( AP )} \\
    \midrule
     A, B &  MILAN, Italy & Italy, MILAN \\
    \midrule
    
    Ent-SpERT & (A,B) & (A,B) \\
    SpERT & (A,B) & (A,B) \\
    TABTO & (A,B) & \textcolor{red}{(B,A)} \\
    PURE & (A,B) & - \\
    \midrule
    
    \midrule
    \textbf{3} & \multicolumn{2}{C{0.8\textwidth}}{In \textbf{A} , downed tree limbs interrupted power in parts of \textbf{B} .} \\
    \midrule
     A, B &  Indianapolis, Indiana & Indiana, Indianapolis \\
    \midrule
    
    Ent-SpERT & (A,B) & \textcolor{red}{(B,A)} \\
    SpERT & (A,B) & \textcolor{red}{(B,A)} \\
    TABTO & (A,B) & \textcolor{red}{(B,A)} \\
    PURE & (A,B) & \textcolor{red}{(B,A)} \\
    \midrule
    
    \midrule
    \textbf{4} & \multicolumn{2}{C{0.8\textwidth}}{The plane , owned by Bradley First Air , of \textbf{A} , \textbf{B} , was carrying cargo to Montreal for Emery Air Freight Corp. , an air freight courier service with a hub at the Dayton airport .} \\
    \midrule
     A, B &  Ottawa, Canada & Canada, Ottawa \\
    \midrule
    
    Ent-SpERT & (A,B) (Dayton airport, Canada) & (Dayton airport, Ottawa) \\
    SpERT & (A,B) (Dayton airport, Canada) & - \\
    TABTO & (A,B) & (A,B) \\
    PURE & (A,B) & (A,B) \\
    
    \bottomrule
    
    \end{tabularx}
    \caption{Some qualitative examples of models' predictions on original (left column) and swapped (right) CoNLL04 sentences for the ``Located in'' relation. This relation is often simply expressed by an apposition of the head and tail separated by a comma. Predictions of incorrect original triples are in red. These examples are obtained from models trained with the same seed ($s=0$).}
    
    \label{table:qualitative_loc}
    \end{table*}

\end{document}